# Large scale deduplication based on fingerprints

**BIYIHA NLEND Jean Aymar**  aymarnlend@gmail.com
*National Advanced School of Engineering, University of Yaounde I*
*Yaounde, Cameroon*

**MOUKOUOP NGUENA Ibrahim**  imoukouo@gmail.com
*National Advanced School of Engineering, University of Yaounde I*
*Yaounde, Cameroon*

**BOUETOU BOUETOU Thomas**  tbouetou@gmail.com
*National Advanced School of Engineering, University of Yaounde I*
*Yaounde, Cameroon*

## Abstract

In fingerprint-based systems, the size of databases increases considerably with population growth. In developing countries, because of the difficulty in using a central system when enlisting voters, it often happens that several regional voter databases are created and then merged to form a central database. A process is used to remove duplicates and ensure uniqueness by voter. Until now, companies specializing in biometrics use several costly computing servers with algorithms to perform large-scale deduplication based on fingerprints. These algorithms take a considerable time because of their complexity in $O(n^2)$, where n is the size of the database. This article presents an algorithm that can perform this operation in $O(2n)$, with just a computer. It is based on the development of an index obtained using a *5 * 5* matrix performed on each fingerprint. This index makes it possible to build clusters of $O(1)$ in size in order to compare fingerprints. This approach has been evaluated using close to *11 4000* fingerprints, and the results obtained show that this approach allows a penetration rate of less than 1%, an almost $O(1)$ identification, and an $O(n)$ deduplication. A base of *10 000 000* fingerprints can be deduplicated with a just computer in less than two hours, contrary to several days and servers for the usual tools.

## I. Introduction

Biometrics is a global technique that establishes a person's identity by measuring one of their physical characteristics. There are several techniques, some more reliable than others, but unfalsifiable and unique to be representative of a single individual among which we find the biometric fingerprint recognition which is the oldest and most widespread. The systems resulting from this technique are used in many applications pertaining to everyday life, civil and anti-crime domains and can build and manage large databases (millions of records) of populations on a strong identification. Two types of AFIS (Automated Fingerprint Identification System) are identified: Civil AFIS (identity card, driver's license, passport, visa, electoral card) and Police AFIS (suspects or criminals). The size of AFIS databases increases considerably with population growth.

In developing countries, because of the difficulty in using a central system when enlisting voters, it often happens that several regional voter databases are created and then merged to form a central database. A process is used to remove duplicates and ensure uniqueness per voter because it happens very often that there are several identical fingerprint records of the same person in this central database. Similar processes are carried out for centralized databases of national identity cards, passports, driving licenses or visas.

Until now, companies specialized in biometrics use several costly computing servers with algorithms to perform large-scale deduplication based on fingerprints. Applied directly, these algorithms take a considerable amount of time to complete this task. They often use a complexity in $O(n^2)$, where n is the size of the base because they compare each fingerprint with the others in order to avoid duplicate failures. To speed up these algorithms, these companies use scalable cluster architectures, parallel execution architectures of lightweight processes, and multiple computing servers [1] [2]. The need for search algorithms which are both precise in terms of recognition and efficient in terms of computing time in such databases has equally significantly improved. These algorithms try mostly to combine two generally contradictory objectives: *the accuracy of the search*, that is the ability of the algorithm to find in the database, fingerprints that look like that of the search with enough relevance (the rate of true duplicates among detected duplicates), and *computing time*, that is the ability to obtain relevant results in a reasonable amount of time, including in very large databases [3].

The fundamental problem is to develop an algorithm that can achieve this type of deduplication, in an acceptable time and with a low cost hardware infrastructure by reconciling these two contradictory objectives.

The purpose of this article is to propose a deduplication algorithm adapted to very large fingerprints databases that can be robust to latent fingerprints with an average cost in computing time and good accuracy to avoid missing out paired fingerprints. Deduplication will be simplified based on fast, accurate fingerprint identification techniques. The conventional identification scheme is complex and very expensive especially in the case of large biometric databases. The process proposed here aims to reduce complexity and improve performance in terms of computing time and identification rate. This technique will be used as a support to automatically detect duplicates when merging two fingerprint databases.

In the following sections, the approach taken to improve this type of deduplication will be discussed. Section II will make a state of the art on the deduplication of fingerprints. Section III presents the contribution made, the different techniques used to implement the deduplication system. Section IV provides details on the operating results of the proposed system tested with the fingerprint databases. The experiments were conducted with FVC2000, FVC2002, FVC2004, NIST09, NIST14 fingerprint databases. Section V presents the conclusion.

## II. Overview

The deduplication of biometric fingerprint data is a very important step in integrating heterogeneous and critical data. It ensures better quality for data that resulting from this process. Thus, the extraction of information from these will be more precise. This is a step found at the end of a chain. This chain also includes steps for the acquisition of descriptors to characterize the information contained in a fingerprint, classification that reduce the search space of a fingerprint in a specific space and recognition or matching. These first three steps constitute the complete fingerprint classification process. Once these basic steps are done, it is possible to declassify to apply deduplication. Figure 2 illustrates the classification process.

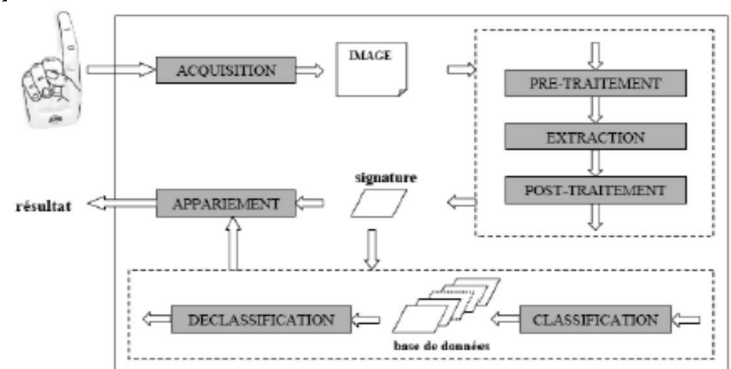

Figure 1 : the complete fingerprint classification process [4]

### II.1. Acquisition of descriptors

The acquisition step of the descriptors consists of fingerprint acquisition, preprocessing and feature extraction with the use of the descriptors. The visual characteristics of fingerprints naturally facilitate the construction of local points of interest. The use of local points of interest in image recognition is a concept that has gradually imposed methods based on points of interest and local descriptors as the most efficient in computing costs and the most effective in terms of comparison, robust to any type and quality of images.

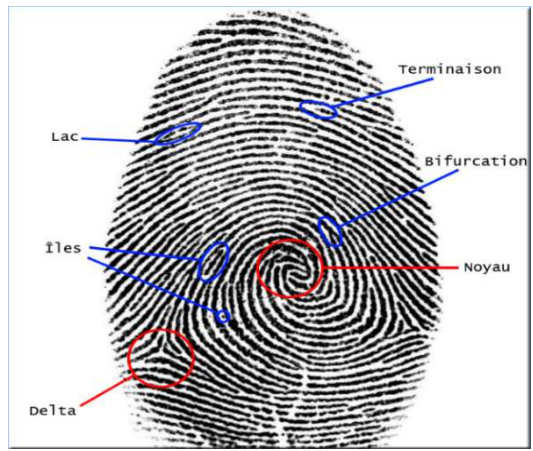

Figure 2 : Minutia localized on a fingerprint [5]

There are mainly two techniques used in **fingerprint acquisition**: *indirect methods (off lines) and direct methods (online)* [6]. Indirect methods are of two types: *rolled type acquisition* with the use of ink and *latent type acquisition* for fingerprints deposited by an individual in public. This last type of acquisition makes it difficult to process the image because the fingerprint is most fuzzy [7]. **Direct methods** use sensors (*acquisition of type poses - slap*). A detail of these sensors is done by Estelle Kiedaisch and al [8].

**Fingerprint preprocessing** is necessary to improve the quality of the image obtained and to avoid errors. *Segmentation (pixel* [9] *or block-based approach* [6]*)* is first used for noise suppression and for two main regions: the bottom region that does not contain a fingerprint and the region of interest (ROI: Region Of Interest) which contains the fingerprint. After segmentation follows the process of filtering the fingerprint in order to improve the quality of the image of the fingerprint (the contrast, the degree of consistency of the lines). The most used algorithms are those developed by Hong and al [10] which are based on the convolution of the image by a Gabor filter and make it possible to preserve the true structure of the lines of the print. This approach consists of the following sub steps: *normalization, ridge orientation estimation, ridge frequency estimation and application of a filtering function, particularly that of Gabor* [11] [12]. Because of latent-type fingerprint acquisition problems, Gabor filter-based fingerprinting algorithms are not very adequate in this regard.

**Feature extraction** makes it possible to collect the main features of a fingerprint: *the minutiae (*bifurcation, lake, island, termination) and the *singular points* (core, delta). There are two methods to extract these features. *The traditional method* that extracts information on a binary skeleton from the filtered image using the *Crossing Number (CN)* of the pixel [11] [13] [14] to determine if it is minutiae; for singular points, the Poincaré method [6] [14] makes it possible to determine them. The direct method is to extract the minutiae directly on the filtered image [15]. In this case, although the direct method is faster, it is inefficient for acquiring noisy prints. The traditional method is preferred and additional processing based on *CN* [4] is necessary to eliminate false minutiae.

Once the extraction is done, a descriptor is used to represent the signature of the fingerprint. It is a representation or set of representations chosen to characterize the information contained in the fingerprint and to facilitate searches. Two approaches of representation can be adopted: the global approach [16] with the use of global descriptors [17] [18] and the local approach with the use of local descriptors [3] [19]. The global approach is not compatible with the latent fingerprint robustness constraint because it uses unreliable or unavailable information (location and orientation of cores or deltas) in this type of fingerprint. The local approach focuses on great robustness to the noise and the occultation of the images. This approach will be given priority over the global approach.

Local fingerprint descriptors can be grouped into five broad categories: *local minutiae structures centered on minutia; local structures of minutiae not centered on minutia; patches of local texture; descriptors based on ridges; hybrid descriptors*. **The local minutiae structures centered on a minutia** have as local reference a central thoroughness and its orientation. Cappelli [20] classifies these structures into two subcategories: *nearest neighbor structures* [21], *and fixed radius structures* such as the "Minutiae Cylinder Code (MCC)" [3] [20] [22]. MCC is rich in local information and allows for better performance without using global consistency. In *local structures of non-centered minutiae* (triplets and n-tuples of minutiae), the triplets of minutiae form another type of descriptor based on neighborhoods of minutiae. They are not defined around a central minutia, but directly as a set of neighboring triplets. The concepts of triplets and tuples of minutiae have been the subject of several works [23] [24] [25] [26] [27]. In *local texture patches* (orientation map patches), the texture of the fingerprint is used as the global fingerprint descriptor, using an absolute landmark such as a core or delta [19] [28] [29]. These descriptors are full of information, and cover a significant area. For **ridge-based descriptors**, if the method of extracting the basic information of the fingerprint image enables their location in a sufficiently precise manner, the ridges may also constitute an interesting base of local descriptors. Several works have been carried out using descriptors based on ridges [30] [31]. Hybrid descriptors are other types of descriptors, not classifiable in the preceding categories or crossing the characteristics of several categories, have also been developed [32] [33]. Due to the constraints of robustness, only comparison methods built on the local descriptors of *minutiae triplet type (local structure of not centered minutiae)* will later be retained.

The information commonly used in descriptors based on minutia triplets are: *length of the sides of the triangle formed by the inter-minutia segments; angles of this triangle; orientation of the triplet minutiae expressed in an intrinsic reference to the triangle*. The use of all possible minutiae triplets for each fingerprint can be problematic. By noting *n,* the number of minutiae present on a fingerprint image, the number of possible triplets changes to $O(n^3)$. It is therefore necessary not to keep a too large number of triplets per image, firstly because the majority of the scores aggregated between local descriptors have computing costs in *O(nb minutiae query x nb minutiae reference)*, then to limit the cost of storing the signature associated with minutiae triplets [3]. Several works have been carried out to restrict minutiae triplets. Since the use of a triangulation of [34] is widespread to restrict the number of triplets retained per image, Xuefeng Liang and al [26] estimated the average number of triplets generated by fingerprint, according to the number of minutiae present.

**II.2. Fingerprint classification**

The information that characterizes a fingerprint is the base of the classification step. It is a capital operation at the base of a deduplication system. It makes it possible to considerably reduce the search space of a query fingerprint in a portion of the database corresponding to the class of this fingerprint. Several fingerprint classification systems have been proposed. Some are based on the general topographic shape of the fingerprint [35], others on support vector machines (SVM) and neural networks [36] [37] to obtain learning rates in the range of 89.1% to 97%. Block ridge let transform associated with a neuro-fuzzy classifier based on the Seugeno model [38] also makes it possible to classify the fingerprint. Local descriptors make it possible to have classifications in more or less fine categories thanks to unsupervised learning techniques, or clustering. These techniques are intended to identify in all local descriptors particular patterns. Ritendra Datta [39] distinguishes three broad categories of clustering algorithms: *matching clustering*, for example spectral clustering, kd-tree [40] [41] [42]; *clustering based on the optimization of a global measure of clustering quality*, for example K-Means [42], Hierarchical K-Means - HKM; *clustering based on statistical modeling* where each cluster is considered as a pattern generated by a random distribution, the set of local descriptors being the combination of these distributions [39]. Other clustering methods are based on the application of a set of hash functions on the data to be clustered, whose objective is to maximize the number of collisions on the signatures or "buckets" of the hash functions points considered heuristically as "close enough". This is the case of Locality-Sensitive Hashing (LSH), introduced by Gionis and al [44].

Once the clustering of local descriptors has been obtained, the overall representation of the image can be constructed as a descriptor distribution or *histogram*. Here again, several methods have been developed to construct these representations (*Hough transform* [45], *Graph representation* [46], *Bag of Words* [47]).

**II.3. Fingerprint recognition**

Fingerprint-based recognition systems have been used primarily for two types of procedures: *verification*, also known as *authentication*, and *identification*. These procedures are also the basis of the deduplication system. In a *verification* procedure, the biometric system is intended to verify whether the identity of an individual characterized by his fingerprints effectively corresponds to the identity he claims. In an *identification* procedure, the biometric system is intended to determine whether an individual, defined by one or more of their fingerprints, belongs or not to a database of reference of individuals also defined by acquisitions of their fingerprints. If so, to identify that individual, providing a list of potential candidates likely to match the individual's query. Several methods exist to compare two fingerprints: ***manual comparison*** [48] and ***automatic comparison***. The first method is long, expensive and accuracy is not guaranteed. In the second method, the computerization of the fingerprint files saves time, improves accuracy in comparisons and archiving a very large number of fingerprints. In this second method, after capturing the digital image come the following operations:

1. **Storing the fingerprint** in an appropriate format. The input format of images to be processed can be made by scanner, digital camera, etc.
2. **Filtering images** by segmentation to eliminate the noise zones of the image (stained, stippled or incomplete image ...).
3. **Evaluating the captured image quality** by calculating factors that help to establish an automatic quality criterion.
4. **Binarization and skeletonization of the image** is necessary in order to have a better detection of the minutiae, to obtain a more schematic image with ridges of the same thickness.
5. **Extracting the minutiae** by a processing software and various algorithms. A data structure is extracted. It is the final process that allows the signature of the fingerprint (digital signature) to be obtained.

To speed up the verification procedures, fingerprint comparison algorithms, capable of automatically providing a comparison score reflecting the matching probability between two fingerprints, have been developed: Jiang and al (JY) [49]; Tico and al (TK) [50]; Parziale and Niel (PN) [51]; Yang and al (QYW) [52]; Medina-Pérez and al (MTK) [53], (MQYW, MJY) [54], (M3gl) [55]. The methods JY and MJY require each time the signature and the skeletonized image to extract local descriptors and to apply the comparison. They are fast when one already has the matrix of the skeleton of the image. The TK, MTK, QYW, MQYW methods use the signature and orientation of the image to extract local descriptors and compare fingerprints. The PN and M3gl methods use only the fingerprint signature to perform the comparison. All these methods use minutia triplets as local descriptors and are latent-based. The PN method is the slowest method. The M3gl method is the fastest. It is important to note that these algorithms have been set up to work for fingerprints with a 500dpi resolution. A resolution conversion is therefore necessary when the image does not have this resolution. Moreover, these methods use for the most part the Euclidean distance to compare two minutiae. Once this score computed, a decision algorithm or a human operator, decide whether the probabilities thus estimated can help to identify the individual in the reference database.

For the identification of a fingerprint, given that a fingerprint is an image and as an image, it can undergo the same search process as that of conventional images. These methods include: *metadata-based methods also known as image metasearch using metadata, content-based image retrieval (CBIR) methods that use image similarity functions, or database-based clustering image data* [39] [56]. However, these methods are not suitable for fingerprint recognition. Indeed, in metasearch, metadata are unavailable or, not very useful for current search scenarios or, in no way contributes to the verification of correspondence between fingerprints. Content-based image search methods require the explicit exhaustive computation of the distances between the totality of the signatures of query images and the totality of the signatures of the reference images in order to calculate a similarity or a dissimilarity between images. This comprehensive work is extremely expensive as databases grow in size, rendering an exhaustive filtering algorithm unusable.

To preserve reasonable computing times, which do not exponentially grow with the size of the referenced databases, the use of non-exhaustive methods (especially those using local descriptors) is essential for the search of images based on fingerprints. These methods comprise of two major steps: *a step to compare local descriptors and a step to check the global consistency of the matchings*. Checking the general consistency is most often obtained by the explicit computation of the optimal similarity optimally superimposing the local descriptors of the request image and the reference image. It can have a significant impact on the comparison time between images, and its use may require particular data structures. Also, according to the order and the nature of the two described stages of local comparison and global consistency, we can classify the comparison algorithms by local descriptors in the following categories:
- **Algorithms using only local matches**. They are usable when the local descriptors are inherently very discriminating between the local patterns actually coming from the same area of the same finger and the local patterns false acceptances or when they are of sufficient size to contain a certain amount of global information [3].
- **Algorithms using an intrinsic geometric orientation of the fingerprint**. They are of two types: the methods using a mark intrinsic to the imprint [57] ; methods that reduce the representation of the fingerprint to a single dimension [58].
- **Algorithms that geometrically align the two sets of local descriptors for each comparison**. It is essentially the generalized Hough transform with its variants, and the Hungarian algorithm [59], [60].
- **The algorithms using a representation by graphs** [61] [62] [63].

Of the various comparison methods, intrinsically oriented methods are inherently insensitive to latent acquisitions, due to the need for global information of sufficient quality. Graph matching methods prove to be too expensive in terms of computation time, and especially not suited to a potential acceleration of the search process in a large database [3]. Only locally matched methods, which can be enriched retrospectively with global consistency information, will be considered, especially those closest neighbors [3] [64]. Indeed, the fingerprint identification algorithms are in essence nearest neighbor search algorithms, and an identification algorithm can be evaluated according to its ability to reveal the fingerprint or paired fingerprints of a fingerprint. query within a database among its closest neighbors. It is therefore a question of being interested in finding nearest neighbors at the local level. The closest neighbor search methods are classified into three main categories [64]: classification trees, hashing methods, and neighborhood graphs. Classification trees include Kd-trees and Hierarchical K-means, and hashing methods include Locality Sensitive Hashing, methods that have also been analyzed in detail by [65].

**II.4. Fingerprint deduplication**

Several works on the elimination of similar data and duplicates exist in the literature. Some works are essentially based on the search for similarities between the texts [66], [67]. Moukouop et al [68] propose several methods for the detection of duplicates in databases whose observed average complexity is less than $O(2n)$ through a simple model highlighting a significant efficiency rate by combining precision and recall. Other work is used in secondary disk storage systems to improve the efficiency of space. This is the case of MAD2 [69], which eliminates duplicate data at the file level and at the block level by using four techniques to speed up the deduplication process and distribute the data in a uniform way: organizing fingerprints in a matrix hash with bucket, fast identification of unduplicated inbound data objects using the Bloom Filter Array (BFA) index, efficient data localization capture and exploitation using Dual Cache, use of a balancing technique for evenly distribute data objects across multiple storage nodes in their backup sequences to further improve performance with a well-balanced load. Subramanian Periyagaram and al [70] use a technique of dividing a block-based dataset into multiple "chunks", with chunk boundaries being independent of block boundaries (due to the hashing algorithm) to organize data to facilitate data deduplication. Benjamin Zhu and al [71] propose a disk data deduplication method based on three techniques which include: (1) the summary vector, a compact data structure in memory for identifying new segments; (2) stream-informed segment formatting, a method of formatting data to improve disk localization for sequentially accessed segments; and (3) location-preserved caching, which maintains fingerprint location of duplicate segments for high cached success rates. Put together, they can remove 99% of disk access for data deduplication. These techniques cannot be directly adapted in the deduplication of fingerprints because it is here images and not texts. However, since fingerprint signatures are text-based, these methods provide solutions that can be adapted to these signatures.

GenKey [72], a company specializing in the Automated Biometric Identity System (ABIS), uses server farms that are capable of handling more than one billion fingerprints per second per server. The deduplication process can be performed continuously and in real time, in parallel with the registration of fingerprints. In this system, duplicates are detected, judged and processed accordingly as soon as they enter the database, resulting in a reliable and always accurate and up-to-date fingerprint database. This system has already proven itself in the management of electoral lists on a large scale in several countries. In Ghana, GenKey made a biometric enrollment of more than 14 million voters in 6 weeks with its ABIS software that ran on 8 standard servers and performed about 980 billion fingerprint comparisons to ensure that every voter was registered once. In Tanzania, more than 23 million voters were registered with 168 million fingerprints in 6 months. In Kenya, 4 million people were registered at mobile points of service for remote locations and deduplication was centralized and in real time. Other projects of this scale have been carried out in Mozambique (9 million voters in 2013), Ghana (15 million voters in 2012, 2015 and 2016), Tanzania (23 million voters in 2015), Cameroon (6.5 million voters in 2013) [73].

Neuro Technology [74], another company specializing in the ABIS system, offers its Mega Matcher solution which is a complete automated biometric identification system for the deployment of large scale multi-biometric projects. Beyond the use of compute farm, it integrates transactions in parallel. This further speeds up the deduplication process. This system has proved its worth in the Democratic Republic of Congo (46.5 million voters with 5.3 million duplicates detected), Venezuela (18 million voters), Bangladesh (80 million voters), Sierra Leone (2.7 million voters). voters), Ukraine (12 million voters), Mexico, Somalia, Bosnia and Herzegovina (4.6 million voters), El Salvador, Angola, Colombia.

Other companies like Ipsidy [75], Gemalto [76], use the same principle of several calculation servers to perform the fast deduplication.

The international company M2SYS uses a Bio-Hyper server ™ [1] system to make superscalar biometric matching. This system integrates biometric software corresponding to a high performance running on HP server hardware with AMD multi-core. This system can exploit up to 64 kernels in a single server to match 100 million fingerprints per second and rapidly de-duplicate millions of stored biometric templates. This system uses a scalable cluster architecture, supports multiple biometric modalities (fingerprint, venal impression, palm footprint and iris). Its server is capable of including four, 12 AMD Opteron core. This system has proven itself in several countries including turkey, Yemen in 2014, Nigeria, the United States of America, Iraq, India, Saudi Arabia.

Although these systems are ultra-fast, they mostly require the use of multiple computing servers to increase their performance. This has an impact on the cost of acquisition and maintenance. Faced with the difficulty of acquiring large computing servers, it is necessary to find an approach that draws resources from classification and identification to reduce the cost of deduplication. The proposed approach will first reduce the cost of identification by substantially reducing it to a complexity in *O (1)* for application to this context.

### III. Contribution

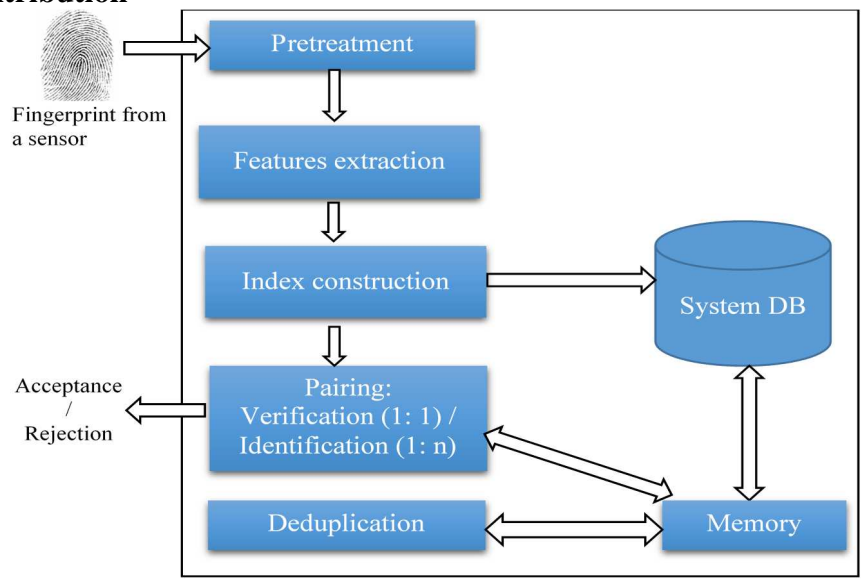

Figure 3 : Recognition and deduplication system

The design of the deduplication system is projected on two different levels: *the design of the system and the design of the algorithms implemented in the system*. The design of the system refers to the different stages of system realization. The design of the algorithms implemented in the system directly hints at the different techniques used for implementation. This section is entirely devoted to the presentation of the deduplication system. The diagram in figure 3 shows the overall operation of the system.

### III.1. Working hypotheses

**Hypothesis 1.**

The step of acquiring local descriptors based on triplets of minutiae to characterize the information contained in a fingerprint is considered as a working basis. This step integrates the following modules: *enrollment, pre-processing, extraction of characteristics and storage of the signature file of the fingerprint*. The following algorithm was built to achieve this signature:
1. Convert the input image to get an image with a resolution of 500dpi;
2. Segment the image to determine the ROI and apply the filtering algorithms. Determine the orientation of the resolution image;
3. Obtain the skeletonized image from the orientation and binarization obtained;
4. Extract the minutiae after removing the false minutiae.

**The enrollment module** that biometrically records individuals in the system's database. During the enlistment phase, the biometric characteristic of an individual is captured by a biometric reader to produce a digital representation of that biometrics. The best way to acquire fingerprints is the use of a sensor (optical, thermal ...). However, as part of this work, the *ink* was used as an acquisition technique, this method being easily accessible. For this purpose, after introducing it into the ink, the finger is printed on a piece of paper, which will undergo scanning by scanner. This technique has been used for a long time in early fingerprinting systems and is of paramount importance in the context of research. Indeed: it shows that the use of a biometric system based on fingerprints is possible even in areas where the technology is not evolved or for projects with a low investment cost.

**Pretreatment** which makes use of certain methods of image processing from the state of the art. the method used in this work successively integrates: *segmentation* for the separation of the fingerprint into two regions (the background region and the region of interest); *Sobel filtering* for contour detection; *pixel orientation estimation* for managing the fingerprint classification process and extracting fingerprint characteristics; *the median filter* for noise reduction to improve the results of future treatments while maintaining the contours of the image; *Gaussian blur* to accentuate the fingerprint in order to improve the quality of detail.

<u>*Result*</u>

Figure 4 shows the resulting images of each of the preprocessing operations.

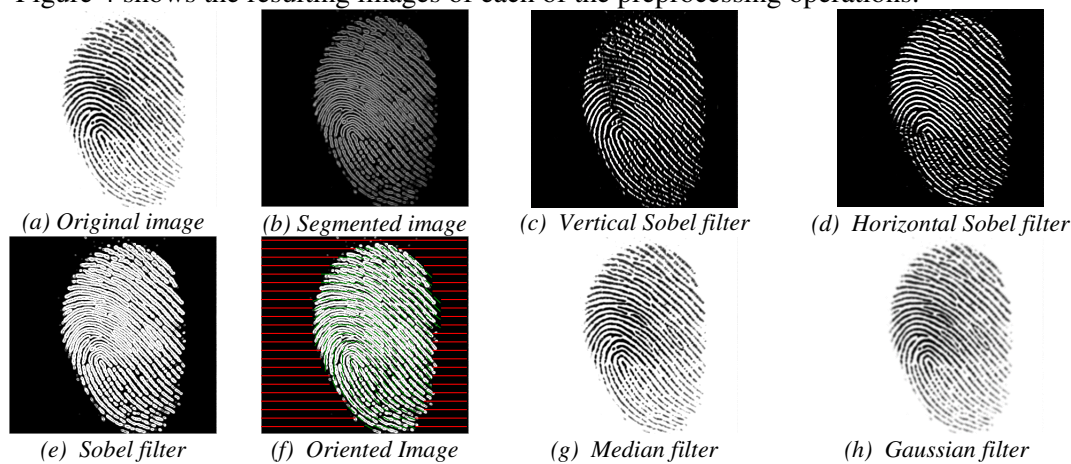

*(a) Original image*  *(b) Segmented image*  *(c) Vertical Sobel filter*  *(d) Horizontal Sobel filter*

*(e) Sobel filter*  *(f) Oriented Image*  *(g) Median filter*  *(h) Gaussian filter*

Figure 4: Result Step Segmentation - Gaussian Filter.

**Feature extraction** that uses the classical method to extract *minutiae* and *singular points*. The image is first prepared in the extraction step by means of binaryization and skeletonization. A *binarization* to obtain an image coded on 2 levels of gray (0 and 1) in order to separate explicitly within the image the zones of peaks and the zones of valleys. A *skeletonization* to reduce the amount of information contained on a line of the fingerprint in order to find the terminal pixels and bifurcations, possibly with an intermediate phase of denoising and correction of the binarized image. Once the skeleton obtained, one easily deduces the position, the orientation of the minutiae and the production of a signature file after the detection, the extraction of the minutiae and the suppression of the false minutiae by applying the Crossing Number (CN) of Galy [21].

*Result*

Figure 5 shows the binary image (5b), the skeletonized image (5c) obtained from the filtered image (5a), the image (5d) after detection of the minutiae, the image (5e) after detection false minutiae and the image (5f) after removing false minutiae.

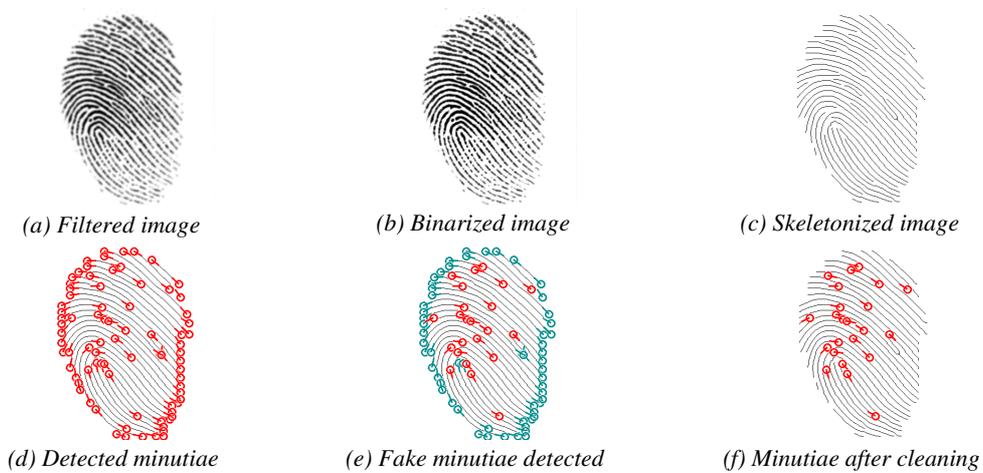

*(a) Filtered image*   *(b) Binarized image*   *(c) Skeletonized image*

*(d) Detected minutiae*   *(e) Fake minutiae detected*   *(f) Minutiae after cleaning*

Figure 5: Result Binarization, skeletonization, detection and suppression of false minutiae

After the minutiae validation, we have a signature file *S* with *N* valid minutiae

$$S = \{M_i = (x_i, y_i, \theta_i, t_i) \mid i \in [1..N]\} \quad \|S\| = N \quad (1)$$

- $(x_i, y_i)$ : the coordinates of the minutia in the image;
- $\theta_i$ : the direction of the local block associated with the streak;
- $t_i$ : type of minutiae: bifurcation or termination.

*Result*

The signature of the fingerprint (Figure 5f), is in the form of Figure 6.

```
207;45;3,33898830413818;1
149;62;0,19739556312561;1
192;51;0,19739556312561;1
245;78;3,75231862068176;1
169;82;3,14159274101257;1
298;88;3,75231862068176;1
172;122;3,33898830413818;1
224;120;3,75231862068176;1
115;142;2,67794513702393;1
164;136;0,0996686518192291;1
177;145;3,43304944038391;1
184;148;0,19739556312561;1
216;166;3,75231862068176;1
264;173;0,785398185253143;1
169;188;3,24126124382019;1
196;181;0,540419518947601;1
148;198;2,41495037078857;1
160;207;0,0996686518192291;1
227;217;3,87440776824951;1
176;229;0,837981224060059;1
146;240;1,27933955192566;1
185;248;1,03037679195404;1
238;328;3,68201208114624;1
```

*Figure 6: Signature of a fingerprint.*

This signature is used for the construction of the index and the classification of the fingerprint.

**Hypothesis 2:**

The comparison method of Medina-Perez Garcia-Borroto et al *(M3gl)* [54] is the one used. It's a quick method that only uses fingerprint signatures to compare

### III.2. Construction de l'index

The working hypothesis considers that a feature vector (signature) per fingerprint has been constructed and is in the form of equation (1). From this signature, the index of this fingerprint is built by applying the following algorithm:

1. Determine the values $x_{min}, y_{min}$ and $x_{max}, y_{max}$ of the minutiae obtained.
2. Make a translation of the abscissae and ordinates of each of the minutiae with respect to $x_{min}$ and $y_{min}$ (for each minutiae, apply $x_i = x_i - x_{min}$; $y_i = y_i - y_{min}$). This translation makes it possible to restrict oneself only to the target zone;
3. Get the new width and height of the target area image from the formulas $l = x_{max} - x_{min} + 1$ and $h = y_{max} - y_{min} + 1$.
4. Form the square matrix $N * N$ of the minutia blocks as follows:
   a. Initialize the matrix to 0 (each block will contain the number of minutiae of the block);
   b. Calculate the width and height of a block by applying the following formulas:
   $$l_{block} = \frac{(x_{max} - x_{min} + 1)}{N} \quad (2)$$
   And
   $$h_{block} = \frac{(y_{max} - y_{min} + 1)}{N} \quad (3)$$
   c. For each translated minutia, calculate its block by applying the following formulas:
   $$x_{block_i} = Whole\ part\left(\frac{x_i}{l_{block}}\right) \quad (4)$$
   And
   $$y_{block_i} = Whole\ part\left(\frac{y_i}{h_{block}}\right) \quad (5)$$
   d. From the obtained coordinates, increment by one the number of minutiae of the corresponding block.
5. The index is obtained by concatenating the numbers of minutiae blocks by separating them by a special character. In this case, the matrix is scanned column by column and the "-" character is used
6. L'index collected defines the digital print categorie

***Result***
The index obtained from the signature of the fingerprint (Figure 6) is in the following form (Figure 7)

1-1-1-1-0-1-4-2-2-0-2-1-2-0-0-1-0-0-1-1-1-0-1-0-0

Figure 7: a digital print index .

**Note: In step 4,**
1. A grouping according to the type of minutiae can be carried out (to distinguish the number of bifurcation and of termination);
2. A determination of the coordinates of the center of the block from the minutiae of the block can also be used;
3. The coordinates of the centers of each block, allow to have the coordinates of the center of gravity of the fingerprint

### III.3. Fingerprint identification

Identification in large databases is expensive in terms of complexity and computational efficiency. Ranking and indexing techniques are needed to limit the number of models that must be compared with the user's signature

#### III.3.1. Data structure in memory

To facilitate identification, indexes were built on each fingerprint to classify them. This resulted in a *penetration rate* of around 1%. The penetration rate is the amount of fingerprints that is transmitted to the filtering algorithm. This identification acts on a hash table having for key the index obtained from the signature of the imprint and for value the list of identifiers (ID) of the fingerprints having the same index. The structure used:

**Hash table (String, List<String>) (TH1)**

This structure will manage collisions. Indeed, in case of collision of the indexes, the identifier of the person is added in the list of identifiers having the same index. The hash function is defined as follows:
```
Function hash (s : string) : integer
var h, i : integer;
begin
   h := 0;
```

```
    for i := 1 to length(s) do
      h=h+Asciicode(s[i]);
    fpour
    return h ;
end
```
  This function can be improved. Some programming languages automatically handle this hash function. The data structure is loaded in O(n). The loading principle is as follows:
1. Retrieve the list of fingerprint indexes and identifiers (IDs) from the database.
2. Browse the list obtained in step 1. For each entry in the list, check if the index exists in the hash table **(TH1)**
   a. If the index exists (we have a collision) then add the current ID in the list of IDs corresponding to this index;
   b. If the index does not exist, add this index in the hash table **(TH1)** as the key of the table and the ID as the first element of the index ID list.

   *III.3.2. Identification algorithm*

  The identification method is as follows:
1. Calculate the index from the signature of the request fingerprint (see paragraph III.2),
2. Locate this index in the hash table **(TH1)**,
3. Get the list of fingerprint IDs with the previous index as key,

  For each ID obtained make a comparison of the signature of this fingerprint and the fingerprint request. If the score obtained is greater than or equal to the threshold score then, this fingerprint is part of the list of fingerprints similar to the request fingerprint.

  **Note**: access to the list of ID fingerprints is in a constant time ($O(1)$). This thanks to the structure used (hash table).

**III.4. Deduplication of fingerprints**

  The principle of deduplication set up within the framework of this article draws its resources in the identification. Since the structure implemented in the context of the identification makes it possible to obtain an identification in $O(1)$, it is used to carry out a deduplication.

  *III.4.1. Data structure in memory*

  Deduplication in place does not directly remove duplicates for later analysis. To facilitate deduplication, a structure composed of a hash table having as key the index of similar fingerprints and a list of string lists has been adopted. The external list contains lists of fingerprints with the same index. These fingerprints are not necessarily similar; they are simply in the same cluster. The internal list has IDs of identical fingerprints and therefore considered duplicates. This structure has the following form:

   **Hash table (String, List<List<String>>) (TH2)**

  This list is based on the algorithm of paragraph III.3.

  *III.4.2. Deduplication algorithm*

  The deduplication principle is as follows:
1. Browse the hash table **(TH1)**, retrieve the current index;
2. For each index retrieved, obtain the list of fingerprints with this index **(List<String>)**;
3. If the size of this list is less than or equal to 1, go to the next index because this list has no duplicates. If not, go to step 4;
4. Get the list of signatures and possibly the list of fingerprint images **(List<object []>)** ;
5. As long as there are still fingerprints in the list obtained previously, apply the following operations:
   a. Remove the fingerprint at the top of the list and add the ID of that fingerprint to a new list. Add this list to the list of IDs **(List<List<String>>)**. This list is initially emptied.
   b. If after the removal of the fingerprint, there are still other fingerprints, compare the fingerprint remove (in progress) with all the remaining fingerprints in the list taking care each time when there is similarity of the add in the list of IDs similar to the fingerprint being identified and remove the fingerprint identifying from the list of fingerprints read through.
6. When the list has been completely exhausted add the result obtained in the hash table **(H2)**.

## IV. Experiences and results

The experiments were carried out with the known fingerprint databases (FVC2000, FVC2002, FVC2004, NIST09, NIST14), fingerprints formed from the NIST09 databases (BD10000, BD20000, BD30000) and NIST14 (BD40000, BD50000). The fingerprints databases formed from other sources of fingerprints (BDAutres: fingerprints acquired manually, and others). The fingerprints database resulting from the union of all previous bases also constituted. The experiments were on a laptop PC core I7-5500U CPU @ 2.4Ghz X 2.4 Ghz RAM: 8GB. The parameters used for this experiment are as follows:

Table 1: Parameters common to experiments.

| Parameters | Value |
|---|---|
| Size of the square matrix | 5 |
| Minimum length between two minutiae (in pixels) | 15 |
| Maximum length between two minutiae (in pixels) | 100 |
| Number of closest neighbors a minutiae | 4 |
| Rate / Matching Score | 90 |
| Number of matched descriptors | 0 |
| Comparison method | M3gl |

The results are shown in the tables below:

Table 2: Result of experiments.

| FBD | Size | Nb class | Avg. | Min P. | Max P. | Std dev | Min P. Rate | Max P. Rate | Duplicates | Duration deduplication (s) |
|---|---|---|---|---|---|---|---|---|---|---|
| FVC2000 | 320 | 320 | 1 | 1 | 1 | 0 | 0,3125% | 0,3125% | 0 | 0,7191 |
| FVC2002 | 320 | 320 | 1 | 1 | 1 | 0 | 0,3125% | 0,3125% | 0 | 0,7411 |
| BDAutres | 1011 | 1009 | 1,002 | 1 | 2 | 0,0632 | 0,0989% | 0,1978% | 4 | 1,7444 |
| FVC2004 | 4001 | 3991 | 1,0025 | 1 | 7 | 0,1026 | 0,0250% | 0,1750% | 0 | 4,6645 |
| BD10000 | 10000 | 9986 | 1,0014 | 1 | 5 | 0,0566 | 0,0100% | 0,0500% | 2 | 10,7835 |
| BD20000 | 20000 | 19934 | 1,0033 | 1 | 22 | 0,1752 | 0,0050% | 0,1100% | 0 | 21,9237 |
| BD30000 | 30000 | 29823 | 1,0059 | 1 | 38 | 0,2806 | 0,0033% | 0,1267% | 9 | 32,7862 |
| BD40000 | 40000 | 39791 | 1,0053 | 1 | 36 | 0,2591 | 0,0025% | 0,0900% | 12 | 45,0318 |
| BD50000 | 50000 | 49754 | 1,0049 | 1 | 42 | 0,2671 | 0,0020% | 0,0840% | 13 | 57,5517 |
| NIST09 | 54000 | 53713 | 1,0053 | 1 | 46 | 0,2911 | 0,0019% | 0,0852% | 20 | 65,2882 |
| NIST14 | 54000 | 53740 | 1,0048 | 1 | 43 | 0,2697 | 0,0019% | 0,0796% | 15 | 65,5147 |
| BD_GLO | 113609 | 112643 | 1,0086 | 1 | 91 | 0,4167 | 0,0009% | 0,0801% | 749 | 163,2234 |

*Explanations:*

- **Fingerprint database (FBD)**: Name of the database.
- **Size** : height of the data base.
- **Number of classes (Nb class):** detected number of index classes.
- **Average (Avg.)**: Average size of fingerprints by index class.
- **Minimum penetration (Min. P)**: Minimum amount of fingerprints that are passed to the filtering algorithm.
- **Maximum penetration (Max P.)**: Maximum amount of fingerprints that are passed to the filtering algorithm.
- **Standard deviation (Std Dev):** Fingerprint dispersion in different fingerprint classes.
- **Minimum penetration rate (Min P. Rate)**: Minimum amount of fingerprints that are passed to the filtering algorithm relative to the size of the database.
- **Maximum Penetration Rate (Max P. Rate):** The maximum amount of fingerprints that is transmitted to the filtering algorithm in relation to the size of the database
- **Duplicates**: number of duplicate fingerprints detected.

The curves in Figures 8, 9, 10, 11 and 12 illustrate the evolution of some static data as a function of the size of the fingerprint database.

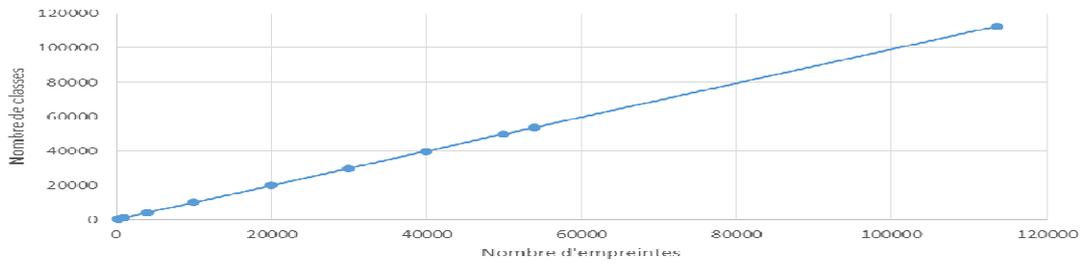
Figure 8 : Curve of evolution of the number of classes

The number of classes changes substantially linearly. Which means that the size of the duplicate prints is relatively small.

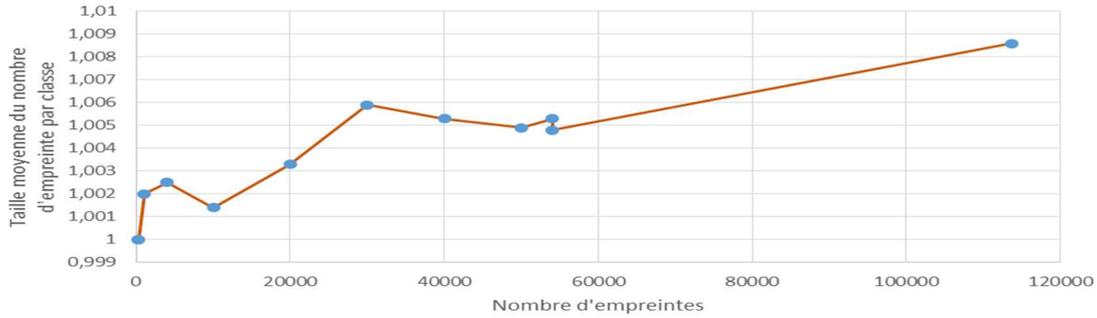
Figure 9 : Average evolution curve by class

The average size of fingerprints per class is relatively small (less than two indexes per class). Which implies that a reduced number of fingerprints is transmitted to a filtering algorithm.

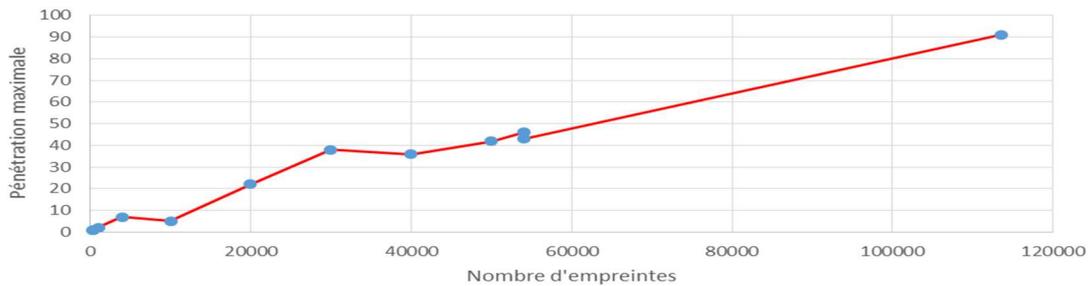
Figure 10 : Evolution curve of maximum class penetration

The maximum penetration changes slightly depending on the size of the database. This implies that the number of fingerprints transmitted to a filtering algorithm increases slightly.

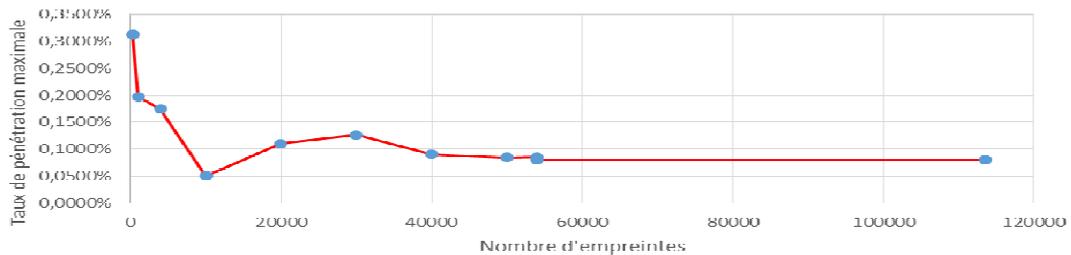
Figure 11 : Evolution curve of the maximum class penetration rate

The penetration rate decreases significantly depending on the size of the database. This implies that the number of fingerprints transmitted to a filtering algorithm increases slightly.

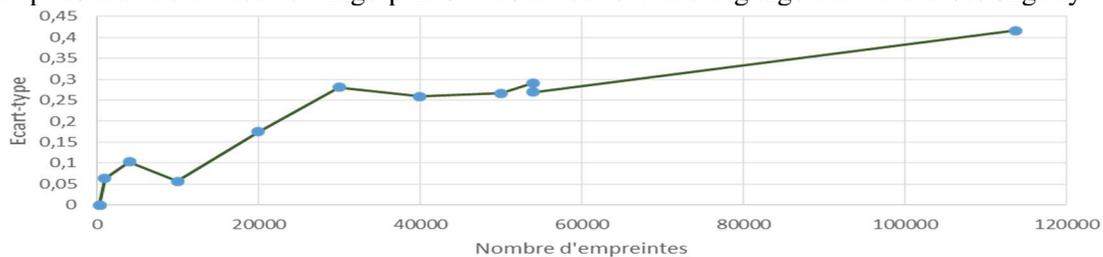
Figure 12 : Evolution curve of standard deviation of class sizes

The standard deviation of class sizes is relatively small. This implies that the dispersion of the fingerprints in the different classes is weak.

These results show that the penetration rate of the database is less than 1% and therefore the size of the fingerprints transmitted to the filtering algorithm is considerably reduced. This result seems better than that produced by Deblonde [3] which announces a penetration rate of 20%.

The existence of a significant linear relationship between two continuous quantitative characters "BD Size" and the "Average Size of Fingerprints by Index Class" made it possible to make a thorough study of this relationship and to formalize the average relationship that unites these two variables using the following equation $Y = aX + b$ (**regression line of Y as a function of X**). Where **Y** is *"Average size of fingerprints per index class"* and **X** is *" BD size"*. This introduces the hypothesis that the values of Y depend on those of X, that is to say postulate that *the knowledge of the values of X makes it possible to predict the values of Y*. The following table presents the basic data for the regression study.

Table 3: Basic data for the study of linear regression.

| BD size | Average size of fingerprints by index class |
|---|---|
| 320 | 1 |
| 1011 | 1,002 |
| 4001 | 1,0025 |
| 10000 | 1,0014 |
| 20000 | 1,0033 |
| 30000 | 1,0059 |
| 40000 | 1,0053 |
| 50000 | 1,0049 |
| 54000 | 1,0053 |
| 113609 | 1,0086 |

Applying the linear regression function on these variables gives results contained in the following table:

Table 4: Results of linear regression.

| Slope (a) | Result (Originally ordered: b) |
|---|---|
| 6,62936E-08 | 1,00177911 |

These results make it possible to estimate the evolution of the *"Average size of fingerprints by index class"* according to *"BD size"*. The results of these estimates are summarized in the table below:

Tableau 5: Evolution of average impressions by class.

| BD size (X) | Average fingerprint size by index class (Y) | Y=aX+b |
|---|---|---|
| 320 | 1 | 1,00180032 |
| 1011 | 1,002 | 1,001846129 |
| 4001 | 1,0025 | 1,002044347 |
| 10000 | 1,0014 | 1,002442043 |
| 20000 | 1,0033 | 1,003104979 |
| 30000 | 1,0059 | 1,003767916 |
| 40000 | 1,0053 | 1,004430852 |
| 50000 | 1,0049 | 1,005093789 |
| 54000 | 1,0053 | 1,005358963 |
| 113609 | 1,0086 | 1,009310661 |
| 7000000 | | 1,465834626 |
| 10000000 | | 1,664715563 |
| 14000000 | | 1,929890146 |
| 20000000 | | 2,327652019 |

Estimate of the number of comparisons for a database of 10,000,000 fingerprints:
- **Average number of impressions per class:** 2
- **Number of comparison operations by class:** (2 (2-1)) / 1 = 1
- **Average number of classes:** 10,000,000 / 2 = 5,000,000
- **Average number of transactions for deduplication:** 5,000,000 * 1 = 5,000,000 comparisons

If a comparison between two fingerprints takes in average 1ms, it will take about **5,000,000ms (1h23'20 ")** to deduplicate this database.

These results show that an electoral file of 10 million fingerprints, can be deduplicated in less than two hours with this approach on a computer with the following features: Intel® Core ™ i5-6400 processor with Intel® HD 530 graphics card (2.7 GHz up to 3.3 GHz, 6 MB cache, 4 cores); DD: 500GB, RAM: 8GB.

By analogy with the previous calculation, the proposed algorithm does fewer fingerprint comparison operations (around 7 million) than the one proposed by GenKey which is 980 billion comparisons for 14 million voters.

## V. Conclusion

The main problem of this article is to develop a deduplication algorithm adapted to very large fingerprint databases, and robust to latent type queries. To implement this algorithm, the local representation approach for fingerprint signatures has been adopted because local representations have the advantage of intrinsic robustness to noisy requests, due to the localized extraction of information. The signature of the fingerprint has been extracted with the following characteristics: the abscissa, the ordinate, the angle and the type of minutia. From the signature obtained, the fingerprint was segmented using a square matrix of size n and several blocks of identical size were obtained. These blocks each contain the number of minutiae in the block. This allowed to build an index and thus to cluster the fingerprints. A hash table built on the obtained index coupled to a list containing the references to the fingerprints made it possible to identify a fingerprint practically in $O(1)$. The search cost is thus limited to the computation of the cluster key, then to the calculations of the distances to the reference descriptors located in the corresponding cluster, resulting in a much reduced total cost compared to a search of the closest ones by exhaustive computation of the distances exact to all descriptors references. This clustering technique served as a support for a sequential algorithm for automatic detection of duplicates when deduplicating a database that could have millions of individuals and to obtain a deduplication substantially in $O(2n)$. The results obtained show performance comparable to the state of the art at low penetration rates, for comparison times slightly lower on a simple computer.

In perspective, it is interesting to look at partial fingerprints, to build an index that can take into account this type of fingerprint and identify the global fingerprint corresponding to this partial fingerprint. It will take for that to build a degree of similarity between the constructed indexes. We can also talk about approximate indexes. In addition, the experiments having been carried out with fixed parameters like the use of a *5\*5* square matrix, it will be interesting to vary these parameters to further evaluate the performance of our contribution by also incorporating fingerprints partial.

## VI. Bibliography

[1] M2SYS Technology, "Multimodal Biometric Matching System for High Speed De-Duplication," 2002-2019. [Online]. Available: http://m2sys.fr/biometric-matching-system/. [Accessed 22 05 2019].

[2] Neurotechnology, "Download biometric algorithm demo software, SDK trials, product brochures," 2015-2019. [Online]. Available: https://download.neurotechnology.com/Neurotec_Biometric_SDK_Documentation.pdf. [Accessed 22 07 2019].

[3] Antoine DEBLONDE, *Algorithmes rapides et fiables de recherche dans une base d'images d'empreintes digitales,* TELECOM ParisTech: Ecole de l'Institut Mines-Télécom, 2015.

[4] Nicolas GALY, *Etude d'un système complet de reconnaissance d'empreintes digitales pour un capteur microsystème à balayage,* Grenoble: Institut National Polytechnique de Grenoble, 2005.

[5] Mahmoud Bentriou, Marion Blanchard, Jimmy Castex and Maxime Chereau, "TPE sur les empreintes digitales !," [Online]. Available: http://empreintesdigitales.free.fr/schema1.png. [Accessed 26 07 2016].

[6] Dario Maio and Davide Maltoni, HandBook of Fingerprint Recognition, second edition ed., Springer, 2009.

[7] Alessandra A. Paulino, Jianjiang Feng and Anil K. Jain, "Latent fingerprint matching using descriptor-based hough transform," *Information Forensics and Security, IEEE Transactions Biometrics Compendium,* vol. 8, no. 1, p. 7, 2013.

[8] Estelle Kiedaisch, Benoit Burg and Lucas Lang, "TPE : Les Empreintes Digitales," [Online]. Available: http://biometrie-tpe68.e-monsite.com/pages/introduction/les-differentes-familles-de-capteurs.html. [Accessed 26 07 2016].

[9] Asker M. Bazen and Sabih H. Gerez, "Segmentation of fingerprint images," *International Journal of Innovative Computing, Information and Control,* p. 6, 2001.

[10] Lie Hong Yifei Wan and Anil Jain, ""Fingerprint image enhancement : algorithm and performance evaluation"," *IEEE Transactions on Pattern Analysis and Machine Intelligence,* vol. 20, no. 8:777, p. 31, 1998.

[11] Raymond Thai, *Fingerprint Image Enhancement and Minutiae Extraction,* Western Australia: The University of Western Australia, School of Computer Science and Software Engineering, 2003.

[12] Anil K. Jain, Soweon Yoon and Jianjiang Feng, "On latent fingerprint enhancement," *Journal of the Institution of Electrical Engineers – Part III : Radio and Communication Engineering,* vol. 93, no. 26, p. 10, novembre 2007.

[13] Kayani Mali and Samayita Bhattacharya, "Various aspect of minutiae as a fingerprint feature," *Proceeding in International Journal of Computer Science and Enginnering Technology,* vol. 1, 2002.

[14] Qinzhi Zhang, Kai Huang and Hong Yan, "Fingerprint classification based on extraction and analysis of singularities and pseudoridges," in *Conferences in Research and Practice Information Technology*, 2002.

[15] Dario Maio and Davide Maltoni, "Direct Gray-Scale Minutiae Detection in Fingerprint," *IEEE Transactions on pattern Analysis and Machine Intelligence,* vol. 19, no. 1, pp. 27-39, 1997.

[16] Aude Oliva and Antonio Torralba, ""Building the gist of a scene : The role of global image features in recognition"," *Progress in brain research,* vol. 155, p. 23–36, 2006.

[17] Anil K. Jain, Salil Prabhakar, Lin Hong and Sharath Pankanti, "Filterbank-based fingerprint matching," *IEEE Transactions on Image Processing,* vol. 9, no. 5, p. 846–859, 2000.

[18] Xudong Jiang, Manhua Liu and Alex C Kot, ""Fingerprint retrieval for identification"," *IEEE Transactions on Information Forensics and Security,* vol. 1, no. 4, p. 532–542, 2006.

[19] David G Lowe, "Object recognition from local scale-invariant features," in *The proceedings of the seventh IEEE international conference on Computer vision*, 1999.

[20] Raffaele Cappelli, Matteo Ferrara and Davide Maltoni, "Minutiae cylinder-code : A new representation and matching technique for fingerprint recognition," *IEEE Transactions on Pattern Analysis and Machine Intelligence,* vol. 32, no. 12, p. 2128–2141, 2010.

[21] Xinjian Chen, Jie Tian and Xin Yang, "A new algorithm for distorted fingerprints matching based on normalized fuzzy similarity measure," *IEEE Transactions on Image Processing,* vol. 15, no. 3, p. 767–776, 2006.

[22] Julien Bringer and Vincent Despiegel, "Binary feature vector fingerprint representation from minutiae vicinities," in *IEEE Fourth International Conference on Biometrics : Theory Applications and Systems (BTAS)*, 2010.

[23] Robert S Germain, Andrea Califano and Scott Colville, "Fingerprint matching using transformation parameter clustering," *IEEE Computational Science and Engineering,* vol. 4, no. 4, p. 42–49, 1997.

[24] George Bebis, Taisa Deaconu and Michael Georgiopoulos, "Fingerprint identification using delaunay triangulation," in *International Conference on Information Intelligence and Systems : Proceedings*, Bethesda, Maryland, USA, 1999.

[25] Bir Bhanu and Xuejun Tan, "Fingerprint indexing based on novel features of minutiae triplets," *IEEE Transactions on Pattern Analysis and Machine Intelligence,* vol. 25, no. 5, p. 616–622, 2003.

[26] Xuefeng Liang, Arijit Bishnu and Tetsuo Asano, "A robust fingerprint indexing scheme using minutiae neighborhood structure and low-order delaunay triangles," *IEEE Transactions on Information Forensics and Security,* vol. 2, no. 4, p. 721–733, 2007.

[27] Ogechukwu Iloanusi, Aglika Gyaourova and Arun Ross, "Indexing fingerprints using minutiae quadruplets," in *IEEE Computer Society Conference on Computer Vision and Pattern Recognition Workshops (CVPRW)*, 2011.

[28] Asker M. Bazen, Gerben TB Verwaaijen, Sabih H Gerez, Leo PJ Veelenturf and Berend Jan van der Zwaag, "A correlation-based fingerprint verification system," *University of Twente, Department of Electrical Engineering, Laboratory of Signals and Systems,* pp. 205-213, 2000.

[29] Xin Shuai, Chao Zhang and Pengwei Hao, "Fingerprint indexing based on composite set of reduced sift features," in *19th International Conference on Pattern Recognition, ICPR*, 2008.

[30] Anil Jain, Lin Hong and Ruud Bolle, "On-line fingerprint verification," *IEEE Transactions on Pattern Analysis and Machine Intelligence,* vol. 19, no. 4, p. 302–314, 1997.

[31] Xiaohui Xie, Fei Su and Anni Cai, "Ridge-based fingerprint recognition," *Advances in Biometrics, Springer,* p. 273–279, 2005.

[32] Cristiano Carvalho and Hani Yehia, "Fingerprint alignment using line segments," *Biometric Authentication, Springer,* p. 380–386, 2004.

[33] Alejandro Chau Chau and Carlos Pon Soto, "Hybrid algorithm for fingerprint matching using delaunay triangulation and local binary patterns," *Progress in Pattern Recognition, Image Analysis, Computer Vision, and Applications, Springer,* p. 692–700, 2011.

[34] Boris Delaunay, "Sur la sphère vide," *Izvestia Akademii Nauk SSSR, Otdelenie Matematicheskikh i Estestvennykh Nauk,* pp. 7:793-800, 1934.

[35] E.R. Henry, *Classification and uses of finger prints,* London, Brodway, Ludgate Hill: George Routledge and Sons Limited, 1900.

[36] Yuan Yao, Gian Luca Marcialis, Massibmiliano Pontil, Paolo Frasconi and Fabio Roli, "A new machine learning approach to fingerprint classification," *Lecture Notes in Computer Science,* p. 2175, 2001.

[37] Paolo Frasconi and Yuan Yao, "Fingerprint classification with combinations of support vector machines," *Lecture Notes in Computer Science,* vol. 1, no. 3, 2012.

[38] M. Tico and P. Kuosmanen, "Neuro-Fuzzy Modeling and Control," *Proceedings of the IEEE,* vol. 83, no. 3, pp. 378-405, 1995.

[39] Ritendra Datta, Dhiraj Joshi, Jia Li and James Z Wang, "Image retrieval : Ideas, influences, and trends of the new age," *ACM Computing Surveys (CSUR),* vol. 40, no. 5, p. 5, 2008.

[40] Jerome H Friedman, Jon Louis Bentley and Raphael Ari Finkel, "An algorithm for finding best matches in logarithmic expected time," *ACM Transactions on Mathematical Software (TOMS),* vol. 3, no. 3, p. 209–226, 1977.

[41] David G Lowe, "Distinctive image features from scale-invariant keypoints," *International journal of computer vision,* vol. 60, no. 2, p. 91–110, 2004.

[42] RM McCabe and EM Newton, "Data format for the interchange of fingerprint, facial, and other biometric information," in *ANSI/NIST*, 2007.

[43] J. B. MacQueen, "Some Methods for classification and Analysis of Multivariate Observations," in *Proceedings of 5th Berkeley Symposium on Mathematical Statistics and Probability*, 1967.

[44] Aristides Gionis, Piotr Indyk and Rajeev Motwani, "Similarity search in high dimensions via hashing," in *Proceedings of the 25th VLDB Conference*, Edinburgh, Scotland, 1999.

[45] Richard O Duda and Peter E Hart, "Use of the hough transformation to detect lines and curves in pictures," *Communications of the ACM,* vol. 5, no. 1, p. 11–15, 1972.

[46] Stefano Berretti, Alberto Del Bimbo and Enrico Vicario, "Efficient matching and indexing of graph models in content-based retrieval," *IEEE Transactions on Pattern Analysis and Machine Intelligence,* vol. 23, no. 10, p. 1089–1105, 2001.

[47] Y-Lan Boureau, Nicolas Le Roux, Francis Bach, Jean Ponce and Yann LeCun, "Ask the locals : multi-way local pooling for image recognition," in *IEEE International Conference on Computer Vision (ICCV)*, 2011.

[48] Science et Vie Junior, La police scientifique - les vrais experts témoignent, vol. 65, 2006, pp. 12-17.

[49] X. Jiang and W. Y. Yau, "Fingerprint Minutiae Matching Based on the Local and Global Structures," in *15th International Conference on Pattern Recognition*, Barcelona, Spain, 2000.

[50] M. Tico and P. Kuosmanen, "Fingerprint matching using an orientation-based minutiae descriptor," *IEEE Transactions on Pattern Analysis and Machine Intelligence,* vol. 25, pp. 1009-1014, 2003.

[51] G. Parziale and A. Niel, "A fingerprint matching using minutiae triangulation," in *1st International Conference on Biometric Authentication*, Hong Kong, China, 2004.

[52] J. Qi, S. Yang and Y. Wang, "Fingerprint matching combining the global orientation field with minutiae," *Pattern Recognition Letters,* vol. 26, pp. 2424-2430, 2005.

[53] M. A. Medina-Pérez, A. Gutiérrez-Rodríguez and M. Garcîa-Borroto, "Improving Fingerprint Matching Using an Orientation-Based Minutiae Descriptor," *Lecture Notes in Computer Science,* vol. 5856, pp. 121-128, 2009.

[54] M. A. Medina-Pérez, M. Garcîa-Borroto, A. E. Gutierrez-Rodriguez and L. Altamirano-Robles, "Improving the multiple alignments strategy for fingerprint verification," *Lecture Notes in Computer Science,* vol. 7329, 2012.

[55] M. A. Medina-Pérez, M. Garcîa-Borroto, A. E. Gutierrez-Rodriguez and L. Altamirano-Robles, "Improving Fingerprint Verification Using Minutiae Triplets," *Sensors,* vol. 12, p. 3418–3437, 2012.

[56] I. Felci Rajam and S. Valli, "A survey on content based image retrieval," *Life Science Journal,* vol. 10, no. 2, p. 2475–2487, 2013.

[57] Joseph H Wegstein, *An automated fingerprint identification system,* US Department of Commerce, National Bureau of Standards, 1982.

[58] Sharat Chikkerur and Nalini Ratha, "Impact of singular point detection on fingerprint matching performance," in *Fourth IEEE Workshop on Automatic Identification Advanced Technologies*, 2005.

[59] Richard O Duda and Peter E Hart, "Use of the hough transformation to detect lines and curves in pictures," *Communications of the ACM,* vol. 15, no. 1, p. 11–15, 1972.

[60] Dana H Ballard, "Generalizing the hough transform to detect arbitrary shapes," *Pattern recognition,* vol. 13, no. 2, p. 111–122, 1981.

[61] DK Isenor and Safwat G Zaky, "Fingerprint identification using graph matching," *Pattern Recognition,* vol. 19, no. 2, p. 113–122, 1986.

[62] Euripides G. M. Petrakis and Christos Faloutsos, "Similarity searching in medical image databases," *IEEE Transactions on Knowledge and Data Engineering,* vol. 9, no. 3, p. 435–447, 1997.

[63] Stefano Berretti, Alberto Del Bimbo and Enrico Vicario, "Efficient matching and indexing of graph models in content-based retrieval," *IEEE Transactions on Pattern Analysis and Machine Intelligence,* vol. 23, no. 10, p. 1089–1105, 2001.

[64] Marius Muja and David Lowe, *Scalable nearest neighbour algorithms for high dimensional data,* 2014.

[65] Mohamed Aly, Mario Munich and Pietro Perona, "Indexing in large scale image collections : Scaling properties and benchmark," *IEEE Workshop on Applications of Computer Vision (WACV),* p. 418–425, 2011.

[66] O. Benjalloun, H. Garcia-Molina, D. Menestria, Q. Su, S. Whang and S. E. J. Widom, "Swoosh: A Generic Approach to Entity Resolution," *The International Journal of Very Large Data Bases (VLDB '09),* pp. 255-276, 2009.

[67] Faouzi Boufarès, Aïcha Ben Salem and Sebastiao Correia, *Un algorithme de déduplication pour les bases et entrepôts de données,* Paris: Laboratoire LIPN - UMR 7030 – CNRS, Université Paris 13, 2012, p. 8.

[68] Moukouop Nguena Ibrahim and Amolo-Makama Ophélie Carmen Richeline, "Fast Semantic Duplicate Detection Techniques in Databases," *Journal of Software Engineering and Applications,* no. 10, pp. 529-545, 2017.

[69] Jiansheng Wei, Hong Jiang, Ke Zhou and Dan Feng, "MAD2: A scalable high-throughput exact deduplication approach for network backup services," in *IEEE 26th Symposium on Mass Storage Systems and Technologies (MSST)*, Incline Village, NV, USA, 2010.

[70] Subramanian Periyagaram, Rahul Khona, Dnyaneshwar Pawar and Sandeep Yadav, "System and method for organizing data to facilitate data deduplication," *Patent Application Publication,* 2010.

[71] Benjamin Zhu, Kai Li and Hugo Patterson, "Avoiding the Disk Bottleneck in the Data Domain Deduplication File System," in *6th USENIX Conference on File and Storage Technologies*, San José, Califormia, USA, 2008.

[72] GenKey, "Automated Biometric Identification System (ABIS). GenKey," GenKey, 2019. [Online]. Available: https://www.genkey.com/wp-content/uploads/2016/12/GenKey-ABIS-eBook-version-2.0-1.pdf. [Accessed 24 03 2019].

[73] GenKey, "L'ABIS de GenKey est conçu pour la déduplication, l'identification et la vérification à petite et grande échelle," [Online]. Available: https://www.genkey.com/wp-content/uploads/2016/12/Elections_ebook_FR-version-2.0.pdf. [Accessed 20 07 2019].

[74] Neurotechnology, "MegaMatcher ABIS most prominent installations and references from customers.html," Neurotechnology, 2019. [Online]. Available: https://www.neurotechnology.com/megamatcher-abis-references.html. [Accessed 26 05 2019].

[75] Ipsidy, "Integrate to our high-speed fingerprint matching solution for enhanced security.," Ipsidy, 2019. [Online]. Available: https://www.ipsidy.com/identification. [Accessed 2019 01 10].

[76] Gemalto, "La biométrie électorale au service du recensement et de l'authentification des électeurs," Gemalto, 2019. [Online]. Available: https://www.gemalto.com/france/gouv/inspiration/biometrie-electorale. [Accessed 2019 03 02].